\documentclass[11pt,a4paper]{article}
\usepackage[hyperref]{emnlp2021}
\usepackage{times}
\usepackage{latexsym}

\usepackage{url}

\usepackage{booktabs}
\usepackage{amsbsy}
\usepackage{footmisc}
\usepackage{amsmath,amsfonts,amsthm,amssymb}
\usepackage{multirow}
\usepackage{graphicx}
\usepackage{thmtools}
\usepackage{thm-restate}
\usepackage{cleveref}
\usepackage{caption}
\usepackage{subcaption}
\usepackage{todonotes}
\usepackage{arydshln}
\usepackage{placeins}
\usepackage{afterpage}

\usepackage[normalem]{ulem}

\usepackage[T1]{fontenc}

\newcommand{\eat}[1]{}

\newcommand{\x}{\mathbf{x}}
\newcommand{\U}{\mathbf{U}}

\newcommand{\h}{\mathbf{h}}
\newcommand{\cc}{\mathbf{c}}
\newcommand{\W}{\mathbf{W}}
\newcommand{\vv}{\mathbf{v}}

\newcommand{\f}{\mathbf{f}}
\newcommand{\rr}{\mathbf{r}}
\newcommand{\bb}{\mathbf{b}}
\newcommand{\real}{\mathbb{R}}
\newcommand{\X}{\mathbf{X}}

\title{When Attention Meets Fast Recurrence:\\Training Language Models with Reduced Compute}

\author{
Tao Lei\\
ASAPP, Inc.\\
{\tt taoleics@gmail.com}\\
}

\date{}

\begin{document}
\maketitle

\begin{abstract}
Large language models have become increasingly difficult to train because of the growing computation time and cost.
In this work, we present SRU++, a highly-efficient architecture that combines fast recurrence and attention for sequence modeling.
SRU++ exhibits strong modeling capacity and training efficiency.
On standard language modeling tasks such as \textsc{enwik8}, \textsc{Wiki-103} and \textsc{billion word} datasets, our model obtains better bits-per-character and perplexity while using 3x-10x less training cost compared to top-performing Transformer models.
For instance, our model achieves a state-of-the-art result on the \textsc{enwik8} dataset using 1.6 days of training on an 8-GPU machine.
We further demonstrate that SRU++ requires minimal attention for near state-of-the-art performance.
Our results suggest jointly leveraging fast recurrence with little attention as a promising direction for accelerating model training and inference.\footnote{Our code, experimental setup and models are available at \url{https://github.com/asappresearch/sru}.}


\end{abstract}
\section{Introduction}
\label{sec:intro}

Many recent advances in language modeling have come from leveraging ever larger datasets and model architectures.
As a result, the associated computation cost for developing such models have grown enormously, requiring hundreds of GPU hours or days per experiment, and raising concerns about the environmental sustainability of current research~\cite{schwartz2020}.
As a consequence, it has become imperative to build \emph{computationally efficient} models that retain top modeling power while reducing computational costs.

The Transformer architecture~\cite{vaswani2017attention} was proposed to accelerate model training and has become the predominant architecture in NLP.
Specifically, it is built entirely upon self-attention and avoids the use of recurrence to enable strong parallelization.
While this change has led to many empirical success and improved computational efficiency, we are interested in revisiting the architectural question:
\emph{\bf Is attention all we need for modeling?}

\begin{figure}[t]
\includegraphics[width=2.95in]{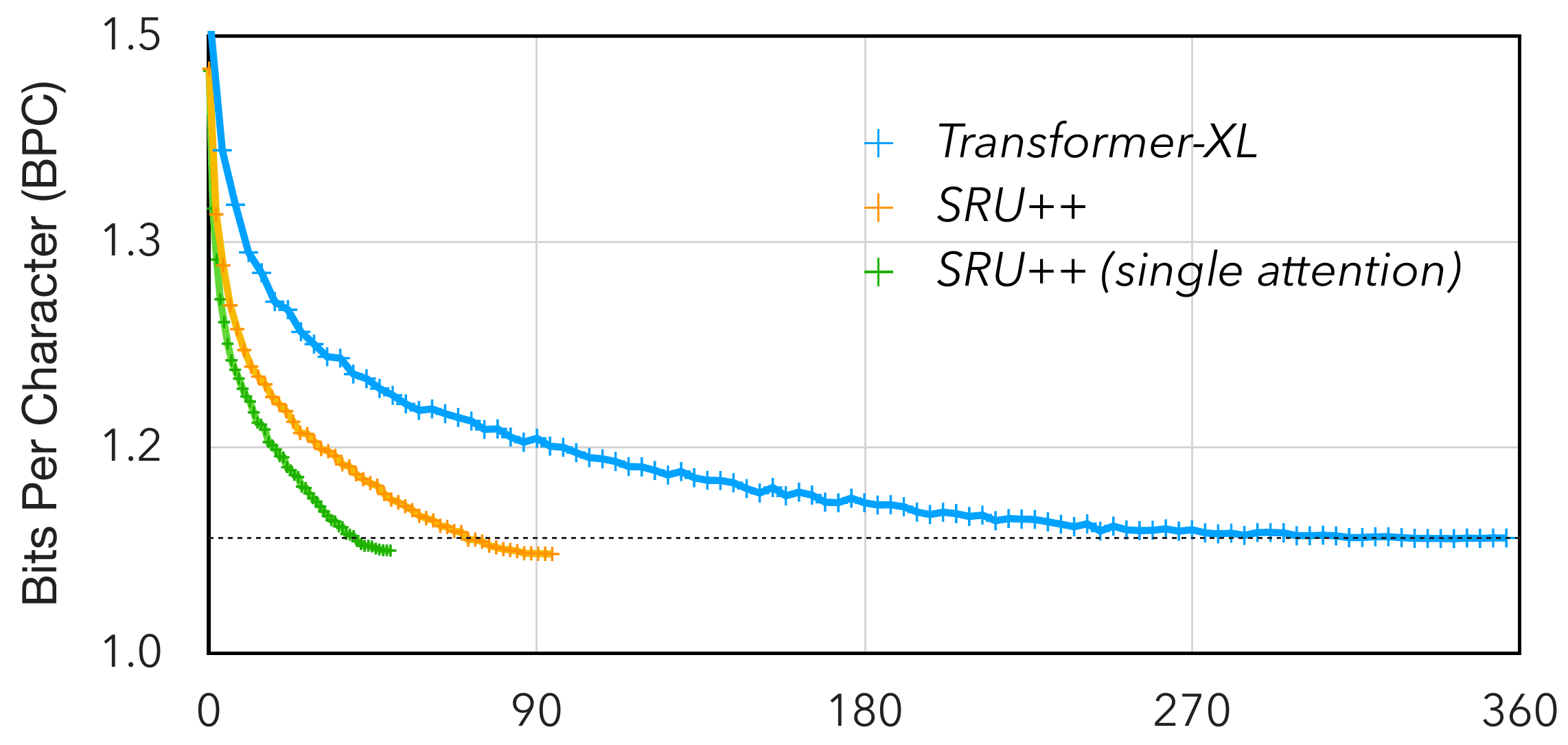}
\caption{Bits-per-character on \textsc{enwik8} dev set vs. GPU hours used for training. SRU++ obtains better BPC by using 1/8 of the resources. 
We compare with Transformer-XL as it is one of the strongest models on the datasets tested.
Models are trained with single precision and comparable training settings.
}
\label{fig:intro}
\end{figure}

The attention mechanism permits learning dependencies between any parts of the input, making it an extremely powerful neural component in many machine learning applications~\cite{Bahdanau:14neuralmt,LinFSYXZB17}.
We hypothesize that this advantage can still be complemented with other computation that is directly designed for sequential modeling.
Indeed, several recent works have studied and confirmed the same hypothesis by leveraging recurrence in conjunction with attention.
For example, \citet{merity2019single} demonstrates that single-headed attention LSTMs can produce results competitive to Transformer models in language modeling.
Other work have incorporated RNNs into Transformer, and obtain better results in machine translation~\cite{lei2018sru,hao-etal-2019-modeling} and language understanding benchmarks~\cite{huang2020transblstm}.
These results highlight one possibility -- we could build more efficient models by combining attention and fast recurrent networks~\cite{bradbury2016quasi,zhang-sennrich-2019-lightweight}.

In this work, we validate this idea and present a self-attentive recurrent unit that achieves strong computational efficiency.
Our work builds upon the SRU~\cite{lei2018sru}, a highly parallelizable RNN implementation that has been shown effective in language and speech applications~\cite{park2018fully,kim-etal-2019-research,hsu-etal-2020-efficient,shangguan2020optimizing}. 
We incorporate attention into the SRU by simply replacing the linear transformation of input with a self-attention component.
The proposed architecture, called SRU++, enjoys enhanced modeling capacity and remains equally parallelizable.
Figure~\ref{fig:intro} compares its performance with the Transformer-XL model~\cite{dai-etal-2019-transformer} on the \textsc{enwik8} dataset.
SRU++ achieves better results while using a fraction of the training resources needed by the baseline.

We evaluate SRU++ on standard language modeling benchmarks including the \textsc{enwik8}, \textsc{Wiki-103} and \textsc{billion word} datasets.
SRU++ consistently outperforms various Transformer models on these datasets, delivering better or on par results while using 3x-10x less computation.
Our model do not use positional encoding, multi-head attention and other techniques useful to Transformer models.
Furthermore, we demonstrate that a couple of attention layers are sufficient for SRU++ to obtain near state-of-the-art performance.
These changes not only highlight the effectiveness of recurrence but also enable strong computation reduction in training and inference.
Finally, we also showcase the effectiveness of SRU++ on the \textsc{IWSLT}'14 De$\rightarrow$En translation task, and open source our implementation in Pytorch to facilitate future research.

\begin{figure*}[!t]
\centering
\includegraphics[width=5.2in]{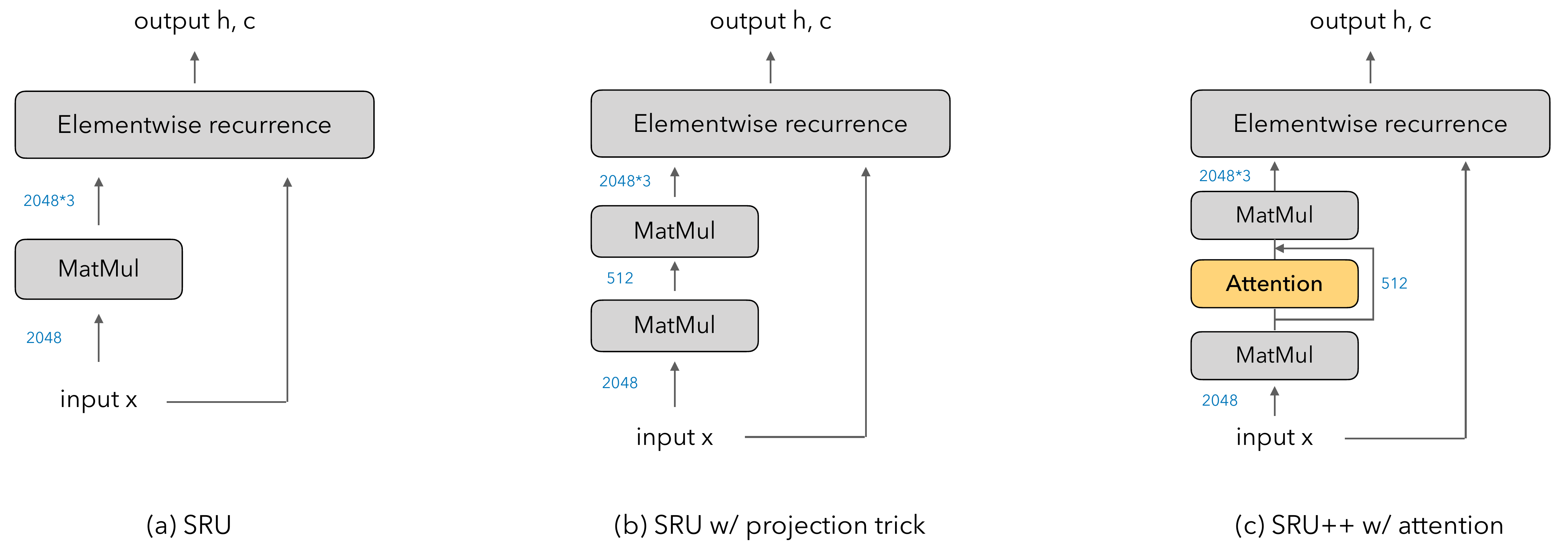}
\caption{An illustration of SRU and SRU++ networks: (a) the original SRU, (b) the SRU variant with projection to reduce the number of parameters, experimented in~\citet{lei2018sru} and (c) SRU++ proposed in this work. Numbers indicate the dimension of intermediate inputs/outputs given hidden size $d=2048$ and attention size $d'=512$.}
\label{fig:arch}
\end{figure*}

\section{Background: SRU}
\label{sec:sru}
We first describe the Simple Recurrent Unit (SRU) in this section. 
A single layer of SRU involves the following computation:
\begin{align*}
\f[t]\ &=\ \sigma\left(\W \x[t] + \vv\odot \cc[t\text{-}1] + \bb\right) \\
\rr[t]\ &=\ \sigma\left(\W'\x[t] + \vv'\odot \cc[t\text{-}1] + \bb'\right) \\
\cc[t]\ &=\ \f[t]\odot \cc[t\text{-}1] + (1-\f[t])\odot (\W'' \x[t]) \\
\h[t]\ &=\ \rr[t]\odot \cc[t]+ (1-\rr[t])\odot \x[t]
\end{align*}
where $\odot$ is the element-wise multiplication, $\W$, $\W'$ and $\W''$ are parameter matrices and $\vv$, $\vv'$, $\bb$ and $\bb'$ are parameter vectors to be learnt during training.
The SRU architecture consists of a light recurrence component 
which successively computes the hidden states $\cc[t]$ by reading the input vector $\x[t]$ for each step $t$.
The computation resembles other gated recurrent networks such as LSTM~\cite{hochreiter1997long} and GRU~\cite{cho-al-emnlp14}.
Specifically, the state vector $\cc[t]$ is a weighted average between the previous state $\cc[t\text{-}1]$ and a linear transformation of the input $\W''\x[t]$.
The weighted aggregation is controlled by a forget gate $\f[t]$ which is a sigmoid function over the current input and hidden state.
Once the internal state $\cc[t]$ is produced, SRU uses a highway network to introduce a skip connection and compute the final output state $\h[t]$.
Similarly, the information flow in the highway network is controlled by a reset gate $\rr[t]$.

Two important code-level optimizations are performed to enhance the parallelism and speed of SRU.
First, given the input sequence $\X = \{\x[1],\cdots, \x[L]\}$ where each $\x[t]\in \real^d$ is a $d$-dimensional vector, SRU combines the three matrix multiplications across all time steps as a single multiplication.
This significantly improves the computation intensity (e.g. GPU utilization).
Specifically, the batched multiplication is a linear projection of the input tensor $\X \in \real^{L\times d}$:
\begin{align}
\U^\top \ &=\ \left(\begin{array}{l}\W \\ \W' \\ \W'' \end{array}\right) \X^\top \;\;,
\end{align}
where $\U \in \mathbb{R}^{L\times 3\times d}$ is the output tensor, $L$ is the sequence length and $d$ is the hidden state size.

The second optimization performs all element-wise operations in an efficient way.
This involves
\begin{align}
    \mathbf{f}[t]&=\sigma(\mathbf{U}[t,0] + \mathbf{v}\odot \mathbf{c}[t\text{-}1] + \mathbf{b}) \\
    \mathbf{r}[t]&=\sigma(\mathbf{U}[t,1] + \mathbf{v}'\odot \mathbf{c}[t\text{-}1] + \mathbf{b}') \\
    \mathbf{c}[t] &= \mathbf{f}[t]\odot \mathbf{c}[t\text{-}1] + (1-\mathbf{f}[t])\odot \mathbf{U}[t,2] \\
    \mathbf{h}[t] &= \mathbf{r}[t]\odot \mathbf{c}[t] + (1-\mathbf{r}[t]) \odot \mathbf{x}[t] .
\end{align}
Similar to other built-in operations such as attention and cuDNN LSTM~\cite{cudnnlstm}, SRU implements all these operations as a single CUDA kernel to accelerate computation.
Note that each dimension of the hidden vectors is independent once $\U$ is computed. 
The computation can run in parallel across each hidden dimension (and each input sequence given a mini-batch of multiple sequences).

\section{SRU++}
The key modification of SRU++ is to incorporate more expressive non-linear operations into the recurrent network.
Note that the computation of $\U$ (Equation 1) is a linear transformation of the input sequence $\X$.
We can replace this linear transformation with self-attention operation to enhance modeling capacity.

Specifically, given the input sequence represented as a matrix $\mathbf{X} \in \mathbb{R}^{L\times d}$, the attention component computes the query, key and value representations using the following multiplications,
\begin{align*}
    \mathbf{Q} &= \W^q\, \X^\top\\
    \mathbf{K} &= \W^k\, \mathbf{Q}\\
    \mathbf{V} &= \W^v\, \mathbf{Q}
\end{align*}
where $\mathbf{W}^q \in \real^{d'\times d}$, $\mathbf{W}^k, \mathbf{W}^v \in \real^{d'\times d'}$ are model parameters.
$d'$ is the attention dimension that is typically much smaller than $d$.
Note that the keys $\mathbf{K}$ and values $\mathbf{V}$ are computed using $\mathbf{Q}$ instead of $\mathbf{X}$ such that the weight matrices $\mathbf{W}^k$ and $\mathbf{W}^v$ are significantly smaller.
We also tested another variant in which we first project $\mathbf{X'} = \mathbf{W}\mathbf{X}^\top$ into the lower dimension $d'$, and then apply three independent $d'$-by-$d'$ matrix multiplications over $\mathbf{X'}$ to obtain the query, key and value representations. This variant achieves similar results.

Next, we compute a weighted average output $\mathbf{A} \in \real^{d'\times L}$ using the scaled dot-product attention introduced in~\citet{vaswani2017attention},
\begin{align*}
\mathbf{A}^\top = \text{softmax}\left(\frac{\mathbf{Q}^\top \mathbf{K}}{\sqrt{d'}}\right) \mathbf{V}^\top.
\end{align*}
The final output $\U$ required by the elementwise recurrence is obtained by another linear projection,
\begin{align*}
\U^\top = \mathbf{W}^o \left(\mathbf{Q} + \alpha\cdot\mathbf{A}\right).
\end{align*}
where $\alpha\in \real$ is a learned scalar and $\mathbf{W}_o \in \mathbb{R}^{3d\times d'}$ is a parameter matrix.
$\mathbf{Q} + \alpha\cdot\mathbf{A}$ is a residual connection which improves gradient propagation and stabilizes training.
We initialize $\alpha$ to zero and as a result,
\begin{align*}
\U^\top = \mathbf{W}^o\, \mathbf{Q} = (\mathbf{W}^o\, \mathbf{W}^q)\, \X^\top
\end{align*}
initially falls back to a linear transformation of the input $\X$ skipping the attention transformation.
Intuitively, skipping attention encourages leveraging recurrence to capture sequential patterns during early stage of training. As $|\alpha|$ grows, the attention mechanism can learn long-range dependencies for the model.
In addition, $\mathbf{W}^o \mathbf{W}^q$ can be interpreted as applying a matrix factorization trick with a small inner dimension $d' < d$, reducing the total number of parameters.
Figure~\ref{fig:arch} (a)-(c) compares the differences of SRU, SRU with this factorization trick (but without attention), and SRU++ proposed in this section.



The last modification is adding layer normalization~\cite{ba2016layer} to each SRU++ layer. 
In our implementation, we apply normalization after the attention operation and before the matrix multiplication with $\mathbf{W}^o$,
\begin{align*}
\U^\top = \W^o\,\,\text{layernorm}(\mathbf{Q} + \alpha\cdot\mathbf{A}) .
\end{align*}
This implementation is post-layer normalization in which the normalization is added after the residual connection.
Alternatively, pre-layer normalization~\cite{pmlr-v119-xiong20b} only applies to the non-linear transformation.
While pre-normalization tends to be less sensitive to different learning rates, we use post-normalization for better results following the observations in~\citet{liu-etal-2020-understanding}.
We analyze the effectiveness of layer normalization in Appendix~\ref{sec:appendix:layer_norm}.

\section{Experimental setup}
\label{sec:setup}

\paragraph{Datasets}
We evaluate our model on four standard NLP benchmarks.
\begin{itemize}
    \item \textsc{Enwik8}~\cite{hutter2012human} is a character-level language modeling dataset consisting of 100M tokens taken from Wikipedia. 
    The vocabulary size of this dataset about 200.
    We use the standard 90M/5M/5M splits as the training, dev and test sets, and report bits-per-character (BPC) as the evaluation metric.
    \item \textsc{Wiki-103}~\cite{merity2016pointer} is a word-level language modeling dataset. The training data contains 100M tokens extracted from Wikipedia articles. 
    Following prior work, we use a vocabulary of 260K tokens, and adaptive embedding and softmax layers~\cite{grave2017efficient,baevski2018adaptive}.
    \item \textsc{Billion word}~\cite{billionword} is one of the largest language modeling datasets containing 768M tokens for training.
    Unlike \textsc{Wiki-103} in which sentences in the same article are treated as consecutive inputs to model long context, the sentences in \textsc{billion word} are randomly shuffled.
    Following \citet{baevski2018adaptive}, we use a vocabulary of 800K tokens, adaptive embedding and softmax layers. 
    \item \textsc{IWSLT}'14 De$\rightarrow$En is a low-resource machine translation dataset consists of 170K translation pairs.
    We showcase SRU++ can be applied to other tasks such as translation.
    We follow the same setup of \citet{lin-etal-2020-autoregressive} and other previous work.
    The dataset uses a shared vocabulary of 14K BPE tokens.
\end{itemize}

\begin{table}[t!]
    \centering
    \begin{tabular}{lcr}
        \toprule
        \bf Model & \bf Batch size $B\times M$ & \bf BPC $\downarrow$ \\
        \hline
        Trans-XL & 24$\times$512 & 1.06 \\
        SRU++ &  24$\times$512 & 1.03 \\
        SRU++ &  16$\times$768 & 1.02 \\
        \bottomrule
    \end{tabular}
    \caption{Test BPC of SRU++ and Transformer-XL on \textsc{Enwik8} dataset. We train SRU++ using the same setting as Transformer-XL base model. 
    Numbers are smaller the better.
    $B$ is the number of sequence. $M$ is the unroll size (and additional context size).}
    \label{tab:enwik8_base}
\end{table}

\paragraph{Models}
All our language models are constructed with a word embedding layer, multiple layers of SRU++ and an output linear layer followed by softmax operation. 
We use single-head attention in each layer and 10 SRU++ layers for all our models.
We use the same dropout probability for all layers and tune this value according to the model size and the results on the dev set.
By default, we set the hidden dimension $d : d' = 4 : 1$.
We report additional analysis and tune this ratio for best results in Section~\ref{sec:results} and Appendix~\ref{sec:appendix:more_analyses}.

For simplicity, SRU++ does not use recent techniques that are shown useful to Transformer such as multi-head attention, compressed memory~\cite{Rae2020Compressive}, relative position~\cite{shaw-etal-2018-self,press2020shortformer}, nearest-neighbor interpolation~\cite{Khandelwal2020Generalization} and attention variants to handle very long context~\cite{sukhbaatar-etal-2019-adaptive,roy2020efficient}.

We compare with previous Transformer models that incorporate one or several these techniques.
However, we do not compare with results that use additional data or dynamic evaluation~\cite{Graves13,pmlr-v80-krause18a},
for a fair comparison between all models.

\begin{table}[!t]
    \centering
    \begin{tabular}{lccr}
    \toprule
    \bf Model & \bf Param & \bf BPC $\downarrow$ & \bf GPU hrs$\,\downarrow$\\
    \hline
    Trans-XL & 41M & 1.06 & 356$\ \,$\\
    SHA-LSTM & 54M & 1.07 & 28$^\dagger$\\
    \hline
    $k=1$ & \multirow{5}{*}{42M} & 1.022 & 37$^\dagger$\\
    $k=2$ & & 1.025 & 29$^\dagger$\\
    $k=5$ & & 1.032 & 24$^\dagger$\\
    $k=10$ & & 1.033 & 22$^\dagger$\\
    No attention & & 1.190 & 20$^\dagger$\\
    \bottomrule
    \end{tabular}
    \caption{Results of SRU++ on \textsc{enwik8} by enabling attention every $k$ layers. We adjust the hidden size so the number of parameters are comparable. $\dagger$ indicates mixed precision training.}
    \label{table:analyze_attention}
\end{table}

\paragraph{Optimization}
We use RAdam~\cite{liu2019radam} with the default $\beta$ values as our optimizer.
RAdam is a variant of Adam optimizer~\cite{Kingma:14adam} that is reported less sensitive to the choice of learning rate and warmup steps while achieving similar results at the end.
We use a fixed weight decay of 0.1 and an initial learning rate of 0.0003 in our experiments.
These values are selected based on \textsc{enwik8} dev set and used for other tasks.
See Appendix~\ref{sec:appendix:tuning_lr_wd} for more details.
We use a cosine learning rate schedule following~\citet{dai-etal-2019-transformer}.
We do not change the initial learning rate unless otherwise specified.
See Appendix~\ref{sec:appendix:train_details} for the detailed training configuration of each model.

Each training batch contains $B$ sequences (i.e. the batch size) and $M$ consecutive tokens for each sequence (i.e. the unroll size), which gives an effective size of $B\times M$ tokens per batch.
Following standard practice, the previous training batch is provided as additional context for attention, which results in a maximum attention length of $2\times M$.
For \textsc{Enwik8} and \textsc{Wiki-103} datasets, the training data is partitioned into $B$ chunks by concatenating articles and ignoring the boundaries between articles.
For \textsc{billion word} dataset, we follow~\citet{dai-etal-2019-transformer} and concatenate sentences to create the training batches.
Sentences are randomly shuffled and separated by a special token <s> indicating sentence boundaries.

\section{Results}
\label{sec:results}

\paragraph{Does recurrence improve upon attention-only model?}
We first conduct a comparison with the Transformer-XL model~\cite{dai-etal-2019-transformer} on \textsc{enwik8} dataset\footnote{\url{https://github.com/kimiyoung/transformer-xl/tree/master/pytorch}}.
Their base model consists of 41M parameters and 12 Transformer layers.
Following the official instructions, we reproduced the reported test BPC of 1.06 by training with 4 Nvidia 2080 Ti GPUs.
The training took about 4 days or a total of 360 GPU hours equivalently.

\begin{figure}[!t!]
    \includegraphics[width=3.0in]{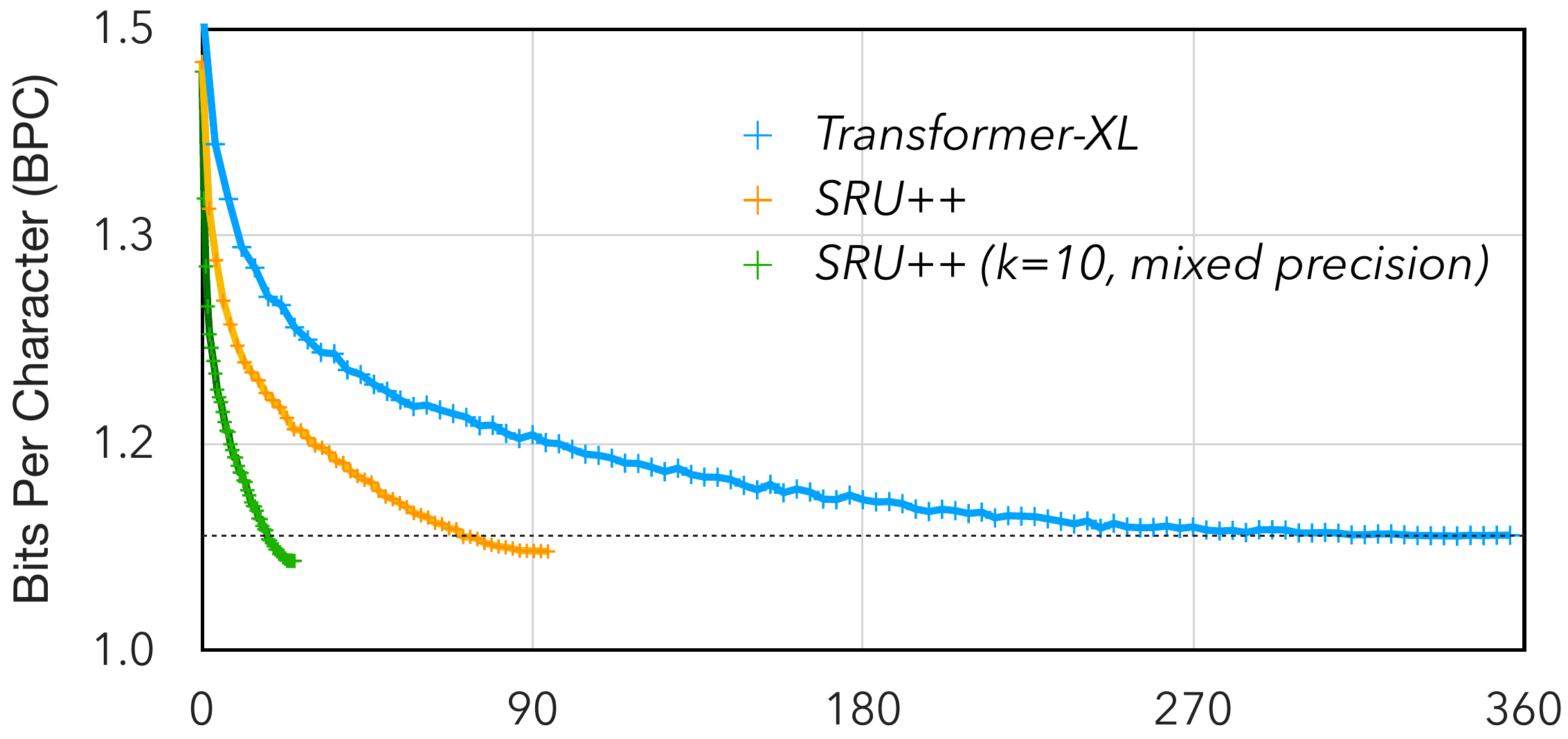}
    \caption{Dev BPC vs. total GPU hours used on \textsc{enwik8} for each model. 
    Using automatic mixed precision (amp) and only one attention sub-layer achieves 16x reduction.
    To compute the dev BPC, the maximum attention length is the same as the unroll size $M$ during training.}
    \label{fig:enwik8_base}
\end{figure}

We train a 10-layer SRU++ model with 42M parameters. For a fair comparison, we use the same hyperparameter setting including the effective batch size, attention context length, learning rate and the number of training iterations as the Transformer-XL base model.
Notably, our base model can be trained using 2 GPUs due to less GPU memory usage. 
After training, we set the attention context length to 2048 for testing, similarly to the Transformer-XL baseline.
Table~\ref{tab:enwik8_base} presents the results.
Our model achieves a test BPC of 1.03, outperforming the baseline by a large margin.
This result suggests that combining recurrence and attention can greatly outperform an attention-only model.
We obtain a BPC of 1.02 by extending the attention context length from 512 to 768, while keeping the number of tokens per batch the same.

\paragraph{How much attention is needed?}
\citet{merity2019single} demonstrated that using a single attention layer with LSTM retains most of the modeling capacity compared to using multiple attention layers.
We conduct a similar analysis to understand how much attention is needed in SRU++.
To do so, we only enable attention every $k$ layers.
The layers without attention become the variant with dimension projection illustrated in Figure~\ref{fig:arch} (b).
Note that $k=1$ gives the default SRU++ model with attention in every layer, and $k=10$ means only the last layer has attention in a 10-layer model.

\begin{figure}[!t!]
\includegraphics[width=3.0in]{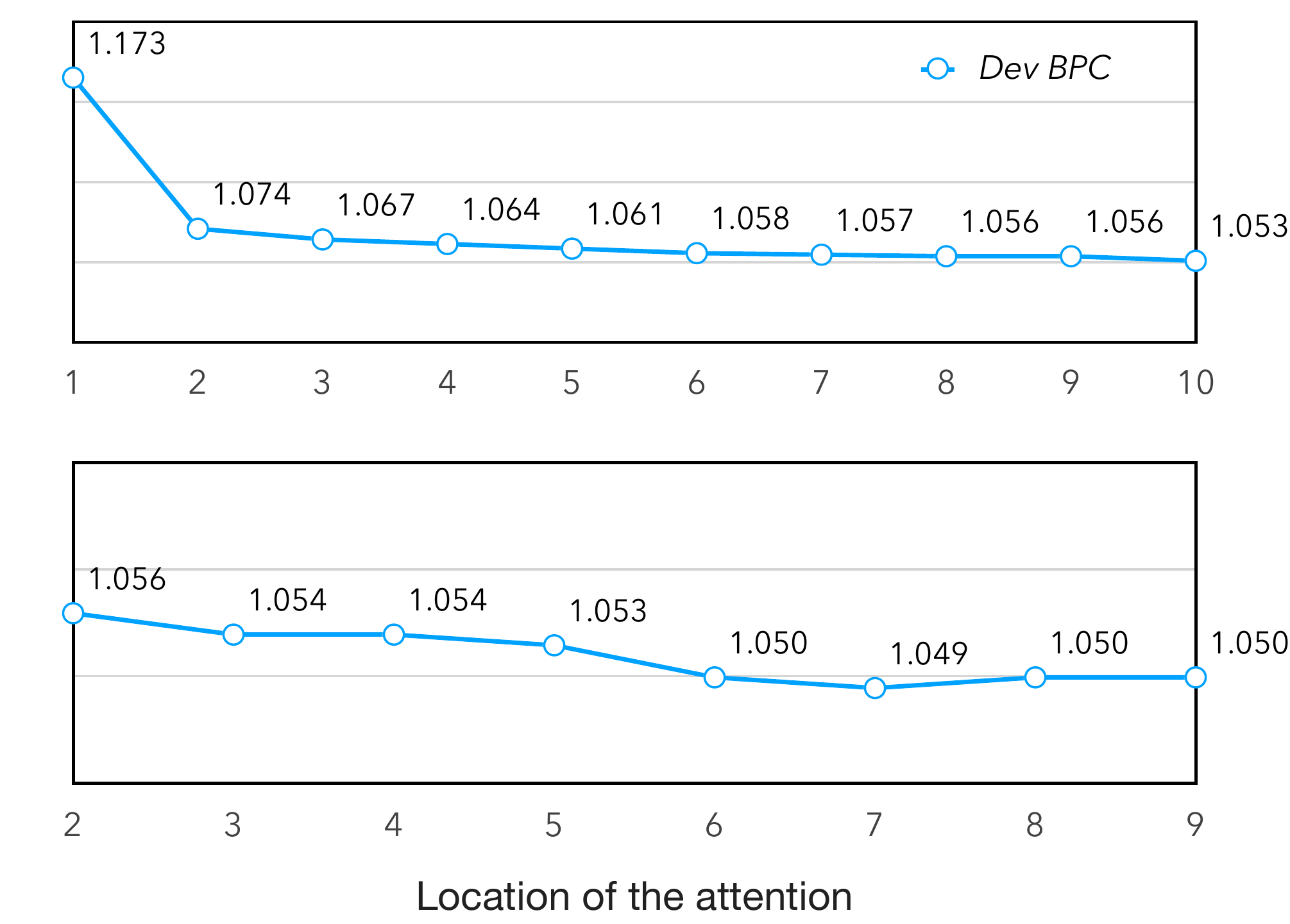}
\caption{
Analyzing where to apply attention.
We enable only one attention layer (top figure) or two (bottom figure) in the SRU++ model.
For the latter, we always apply attention in the last layer and move the location of the other.
X-axis is the layer index. The layer closest to the input embedding layer has index 1.
}
\label{fig:attn_position}
\end{figure}

\begin{table*}[!t]
    \centering
    \begin{tabular}{lccr}
    \toprule
    \bf Model & \bf Parameters $\downarrow$ & \bf Test BPC $\downarrow$ & \bf GPU days $\downarrow$\\
    \hline
    Longformer 30L~\cite{Beltagy2020Longformer} & 102M & 0.99 & 104$^\dagger$\\
    All-attention network 36L~\cite{sukhbaatar2019augmenting} & 114M & 0.98 & 64$\ \,$\\
    Transformer-XL 24L~\cite{dai-etal-2019-transformer}& 277M & 0.99 & - \\
      $\quad\circ$ $\ $ Compressive memory~\cite{Rae2020Compressive} & - & 0.97 & -\\
    Feedback Transformer~\cite{fan2020accessing} & 77M & 0.96 & -\\
    \hline
    SRU++ Base & 108M & 0.97 & 6$^\dagger$ \\
    $\quad\circ$ $\ $ only 2 attention layers ($k=5$) & 98M & 0.98 & 4$^\dagger$ \\
\hdashline
    SRU++ Large & 191M & 0.96 & 12$^\dagger$ \\
    $\quad\circ$ $\ $ $d=8\,d'$ & 195M & 0.95 & 13$^\dagger$\\
    \bottomrule
    \end{tabular}
    \caption{Comparison with top-performing models on \textsc{enwik8} dataset. We include the training cost (measured by the number of GPUs used $\times$ the number of days) if it is reported in the previous work. 
    Our results are obtained using an AWS p3dn instance with 8 V100 GPUs. 
    The reported training time of all-attention network is based on V100 GPUs while the training time of Longformer is based on RTX8000 GPUs (which is about 90\% speed of V100).
    $\dagger$ indicates mixed precision training.}

    \label{tab:enwik8_sota}
\end{table*}

\begin{table}[!t]
    \centering
    \begin{tabular}{lccc}
    \toprule
    \bf Ratio & \multicolumn{2}{@{~~~}c@{~~~~~}}{\bf Dimensions $d$, $\,d'$} & \bf Dev BPC $\downarrow$ \\
    \hline
    4 & ~3072 & 768 & 0.997 \\
    6 & ~3840 & 640 & 0.992 \\
    8 & ~4480 & 560 & 0.991 \\
    10 & ~5040 & 504 & 0.992 \\
    \bottomrule
    \end{tabular}
    \caption{Dev BPC on \textsc{enwik8} by changing the ratio $d:d'$ in the SRU++ model while fixing the number of parameters to 108M.}
    \label{table:analyze_ratio}
\end{table}

Table~\ref{table:analyze_attention} presents the results by varying $k$.
Our base model is the same 10-layer SRU++ model in Table~\ref{tab:enwik8_base}.
We see that using 50\% less attention ($k=2$) achieves almost no increase in test BPC.
Moreover, using only a single attention module ($k=10$) leads to a marginal loss of 0.01 BPC but reduces the training time by 40\%. 
Our results still outperform Transformer-XL model and single-headed attention LSTM~\cite{merity2019single} greatly by 0.03 BPC.
Figure~\ref{fig:enwik8_base} showcases the training efficiency of our model.
SRU++ is 5x faster to reach the dev BPC obtained by the Transformer-XL model.
Furthermore, using automatic mixed precision training and a single attention layer ($k=10$) achieves 16x reduction on training cost.

\paragraph{Where to use attention?}
Next, we analyze if the location of attention in SRU++ makes a non-trivial difference.
Figure~\ref{fig:attn_position} (top) compares the results by enabling attention in only one of the SRU++ layers. 
Applying attention in the first bottom layer achieves significantly worse result. 
We believe this is due to the lack of positional information for attention, since SRU++ does not use positional encoding.
Enabling attention in subsequent layers gives much better and comparable results because recurrence can encode positional information.

Moreover, SRU++ consistently achieves worse results by moving the attention to lower layer closer to the input embedding.
We also enable a second attention layer while fixing the first one in the 10th layer.
The corresponding results are shown in Figure~\ref{fig:attn_position} (bottom).
Similarly, SRU++ achieves worse results if the attention is added to one of the lower layers.
In contrast, results are comparable once the attention is placed in a high-enough layer.
These observations suggest that the model should first learn local features before attention plays a most effective role at capturing long-range dependencies.
More analyses can be found in Appendix~\ref{sec:appendix:more_analyses}.

\paragraph{Does the ratio $d:d'$ matter?}
Transformer models by default use a FFN dimension that is 4 times larger than the attention dimension~\cite{vaswani2017attention}.
We analyze the ratio of recurrence dimension $d$ to attention dimension $d'$ for SRU++.
A small value of $d'$ can reduce the amount of computation and the number of parameters used in attention layers but may limit the modeling capacity.
Table~\ref{table:analyze_ratio} compares the results of using different $d:d'$ ratio given a similar amount of model parameters.
We fix the model size to around 108M and use 10 SRU++ layers. 
Changing this ratio from 4 to a higher value gives better result.
The best dev result is obtained with a ratio of 8.

Given this observation, we report SRU++ result using a default ratio of 4 as well as a ratio of 8 in the subsequent result sections.
This ensures we conduct a comparison that uses a setup similarly to the default of Transformer models, but also showcases stronger results SRU++ can achieve.

\begin{table*}[!t!]
    \centering
    \begin{tabular}{lccr}
    \toprule
    \bf Model & \bf Parameters $\downarrow$ & \bf Test PPL $\downarrow$ & \bf GPU days $\downarrow$\\
    \hline
    All-attention network 36L~\cite{sukhbaatar2019augmenting} & 133M & 20.6 & -\\
    Feedback Transformer~\cite{fan2020accessing} & 139M & 18.2 & 214$\ \,$\\
    Transformer~\cite{baevski2018adaptive} & 247M & 18.7 & 22$^\dagger$\\
    Transformer-XL 18L~\cite{dai-etal-2019-transformer} & 257M & 18.3 & - \\
    $\quad\circ$ $\ $ Compressive memory~\cite{Rae2020Compressive} & - & 17.1 & -\\
    Routing Transformer~\cite{roy2020efficient} & - & 15.8 & -\\
    kNN-LM~\cite{Khandelwal2020Generalization} & - & 15.8 & -\\
    \hline
    SRU++ Base & 148M & 18.3 & 8$^\dagger$ \\
    \hdashline
    SRU++ Large & 232M & 17.4 & 14$^\dagger$ \\
    $\quad\circ$ $\ $ $d=8\,d'$ & 234M & 17.1 & 15$^\dagger$\\   
    $\quad\circ$ $\ $ only 2 attention layers ($k=5$) & 225M & 17.3 & 11$^\dagger$\\   
    \bottomrule
    \end{tabular}
    \caption{Comparison with top-performing models on \textsc{wiki-103} dataset. We include the training cost (measured by the number of GPUs used $\times$ the number of days) if it is reported in the previous work. The reported training costs are based on V100 GPUs. Our results are similarly obtained using an AWS p3dn instance with 8 V100 GPUs. $\dagger$ indicates mixed precision training.}
    \label{tab:wit103_sota}
\end{table*}
\begin{table}[!t]
    \centering
    \begin{tabular}{lccr}
    \toprule
    \bf Model & \bf ~Param~ & \bf ~PPL$\,\downarrow$~ & \bf Days$\,\downarrow$\\
    \hline
    \multirow{3}{*}{Transformer} & \multirow{2}{*}{331M} & 25.6 & 57$^\dagger$\\
    & & 25.2 & 147$^\dagger$\\
    & 465M & 23.9 & 192$^\dagger$ \\
    \hline
    SRU++ & 328M & 25.1 & 36$^\dagger$\\
    SRU++ ($k=5$) & 465M & 23.5 & 63$^\dagger$\\
    \bottomrule
    \end{tabular}
    \caption{Test perplexity and effective GPU days for training of SRU++ models and the Transformer models of~\citet{baevski2018adaptive} on \textsc{billion word} dataset.}
    \label{tab:1blm}
\end{table}

\paragraph{\textsc{enwik8}}
Table~\ref{tab:enwik8_sota} compares our model with other top-performing models on the \textsc{enwik8} dataset.
We train a base model with $d=3072$ and a large model with $d=4096$ using 400K training steps.
The unroll size and attention context length are set to 1024 during training and 3072 during evaluation.
To compare the computation efficiency we report the effective GPU days -- the number of GPUs multiplied by the number of days needed to finish training. 
Our base model achieves better BPC and uses a fraction of the training cost reported in previous work.
Furthermore, our large models achieve a new state-of-the-art result on this dataset, reaching a test BPC of 0.96 when $d=4\,d'$ and 0.95 when $d=8\,d'$. 

\paragraph{\textsc{Wiki-103}}
Table~\ref{tab:wit103_sota} presents the result of SRU++ models and other top results on the \textsc{Wiki-103} dataset. 
We train one base model with 148M parameters and a few large models which contain about 230M parameters.
As shown in the table, our base model obtains a test perplexity of 18.3 using 8 GPU days of training, about 3x reduction compared to the Transformer model in~\citet{baevski2018adaptive} and over 10x reduction compared to Feedback Transformer~\cite{fan2020accessing}.
Again, changing the hidden size ratio to $d=8\,d'$ improves the modeling capacity.
Our big model achieves a test perplexity of 17.1.
The required training cost remains significantly lower.

\paragraph{\textsc{billion word}}
We double our training iterations to 800K and use a learning rate of 0.0002 for the \textsc{billion word} dataset. 
We train a base model using $d=4096$, $d'=1024$ and an effective batch size of 65K tokens per gradient update.
We also train a large model by increasing the hidden size $d$ to 7616 and the batch size to 98K.
In addition, we use only 2 attention layers ($k=5$) for the large model.
Table~\ref{tab:1blm} reports the test perplexity and associated training cost.
Our base and large model obtain a test perplexity of 25.1 and 23.5 respectively, outperforming the Transformer model of~\citet{baevski2018adaptive} given similar model size.
Moreover, SRU++ achieves 3-4x training cost reduction and is trained using 8 GPUs.
In comparison, the Transformer model uses 32 or 64 V100 GPUs.

\begin{table}[!t]
    \centering
    \begin{tabular}{lcc}
    \toprule
    \bf Model & \bf Speed$\uparrow$ & \bf PPL$\downarrow$\\
    \hline
    kNNLM~(\citeauthor{Khandelwal2020Generalization}) & 145 & 15.8\\
    Trans~(\citeauthor{baevski2018adaptive}) & 2.5k & 18.7\\
    Trans-XL~(\citeauthor{dai-etal-2019-transformer}) & 3.2k & 18.3 \\
    Shortformer~(\citeauthor{press2020shortformer}) & 15k & 18.2\\
    \hline
    SRU++ Large & 15k & 17.1\\
    SRU++ Large ($k=5$) & 22k & 17.3\\
    \bottomrule
    \end{tabular}
    \caption{Inference speed (tokens/second) on \textsc{Wiki-103} test set. Results of baselines are taken from~\citet{press2020shortformer}. We use a single V100 GPU, a batch size of 1 and maximum attention length 2560 for consistency.}
    \label{tab:inference_speed}
\end{table}

\paragraph{Inference speed}
Table~\ref{tab:inference_speed} compares the inference speed of SRU++ with other top-performing models on \textsc{wiki-103} test set.
We use a single V100 GPU for inference.
Our large model runs at least 4.5x faster than all baseline models except Shortformer~\cite{press2020shortformer}.
In addition, our model achieves 0.9-1.1 perplexity lower than Shortformer and runs 50\% faster when using 2 attention layers ($k=5$).

\paragraph{\textsc{IWSLT}}
Does SRU++ work well for other tasks?
We study this question by evaluating SRU++ on the \textsc{IWSLT}'14 De$\rightarrow$En translation task.
We use the open-sourced training and evaluation code of \citet{lin-etal-2020-autoregressive}.
The base model is an 8-layer Transformer model containing 20M parameters.
We train SRU++ models using 6 layers and $d=1024$, resulting in similar number of parameters.
We use the original settings such as learning rate and batch size, except that we use RAdam optimizer for consistency and increase the number of training epochs to 50.
Both architectures achieve much higher BLEU scores given more training epochs.\footnote{\citet{lin-etal-2020-autoregressive} reports a test BLEU of 35.2. We obtain 35.9 for the same Transformer model by training longer.}
Table~\ref{tab:iwslt} presents the test results.
Without additional hyperparameter tuning, SRU++ achieves 0.4 BLEU score higher and less training time compared to the Transformer model tuned in~\citet{lin-etal-2020-autoregressive}.

\begin{table}[!t]
    \centering
    \begin{tabular}{lccr}
    \toprule
    \bf Model & \bf ~Param~ & \bf ~BLEU$\,\uparrow$~ & \bf Hrs$\,\downarrow$\\
    \hline
    Transformer & 20.1M & 35.9$\pm$0.1 & 10.5 \\
    SRU++ & 20.4M & 36.3$\pm$0.2 & 8.5\\
    SRU++ ($k=2$) & 19.6M & 36.1$\pm$0.1 & 7.5\\
    \bottomrule
    \end{tabular}
    \caption{Results on \textsc{IWSLT}'14 De$\rightarrow$En test set. We use a beam size of 5. BLEU scores and training time are averaged over 4 independent runs.}
    \label{tab:iwslt}
\end{table}

\paragraph{Why does SRU++ reduce training cost in our experiments?}
Several factors contribute to the computation reduction observed in our experiments. 
First, combining attention and recurrence gives stronger modeling capacity.
As shown in our experiments, SRU++ often achieves comparable results using fewer layers and/or fewer parameters.
The required computation are much lower for shallower and smaller models. 

We also observe higher training efficiency, requiring fewer training steps and smaller training batch compared to several Transformer models.
For example, SRU++ uses a maximum effective batch size of 98K tokens and 800K training steps on the \textsc{billion word} dataset, while the Transformer model in comparison~\cite{baevski2018adaptive} uses 128K tokens and near 1000K steps.
The reduced batch size and gradient updates cut down the training cost.

Finally, model implementation is an important factor for computation saving.
Our implementation is highly efficient for two reasons.
First, the fast recurrence operation of SRU is a reusable module that is already optimized for speed~\cite{lei2018sru}.
Second, since recurrence encodes positional information, we can use simple single-head attention and remove positional encoding.

\begin{figure*}[!t]
    \includegraphics[width=6.1in]{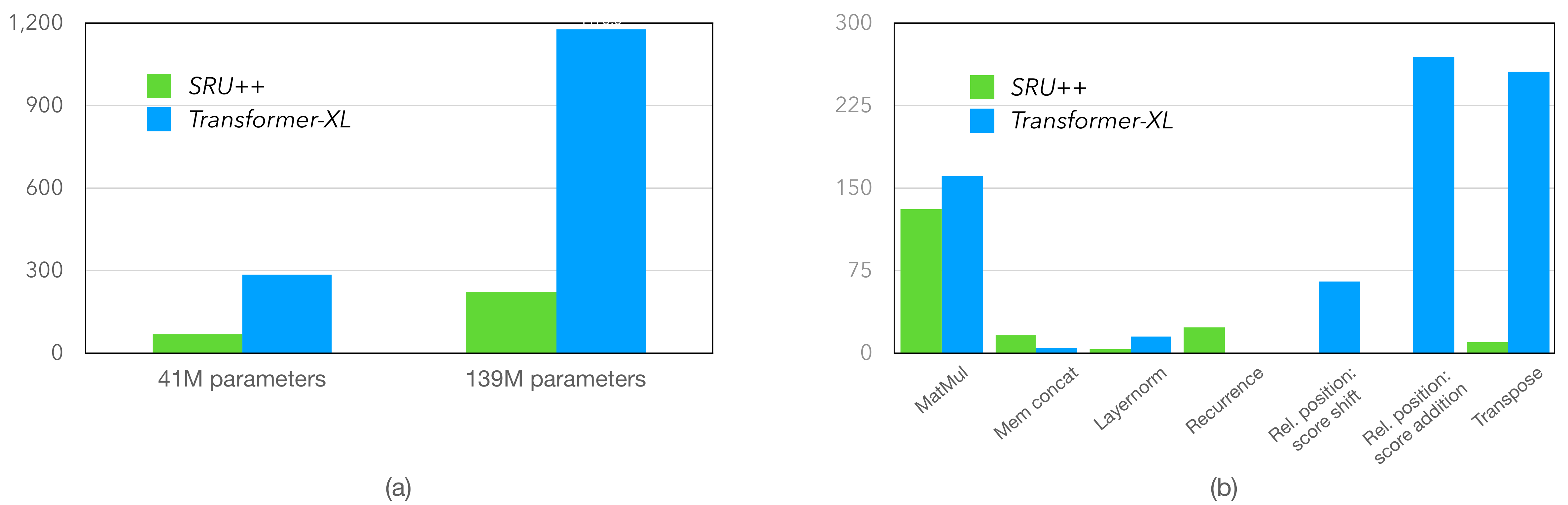}
    \caption{Profiling of SRU++ and Transformer-XL: (a) forward time (in milliseconds) of small and large models and (b) forward time used in various types of time-consuming operations. We use a single GPU for profiling to avoid extra overhead such as data synchronization between GPUs. We use an unroll size / context length $M=512$ and $1024$ respectively for small and large models. All models use a batch size $B=16$ for profiling.}
    \label{fig:profiling}
\end{figure*}

On the contrary, advanced attention and positional encoding mechanism can generate non-trivial computation overhead.
To see this, we measure the running time of SRU++ and Transformer-XL using Pytorch Profiler. 
Figure~\ref{fig:profiling}~(a) shows the average model forward time of a single batch.
SRU++ runs 4-5x times faster compared to the Transformer-XL implementation. 
Figure~\ref{fig:profiling}~(b) breaks down the computation and highlights the most time-consuming operations in both models.
The matrix multiplications are one of the most expensive operations for both models.
Surprisingly, many operations in the relative attention of Transformer-XL are computationally expensive.
For example, the relative attention requires shifting the attention scores and adding up different attention score matrices.
Both require a lot of time but they are not needed in non-relative attention.
In addition, the last column shows the running time of tensor transpose operators needed by batch matrix-matrix multiplications in attention.
Again, the relative attention uses an order of magnitude more time compared to the simple single-head attention used in our model implementation.\footnote{Note that this high latency of tensor transpose might be caused by sub-optimal implementation choices such as a poor arrangement of tensor axes in the open-sourced model. There is room for improvement. Nevertheless, relative attention and positional encoding are reported to be non-trivially slower in other works~\cite{shaw-etal-2018-self,tian2021shatter}.}

\section{Related Work}
\label{sec:related}

Accelerating common architectures for NLP has become an increasingly important research topic recently~\cite{tay2020efficient,sun-etal-2020-mobilebert,Lan2020ALBERT}.
Our work is closely related to two lines of research under this topic.

First, previous works have tackled the speed problem of recurrent neural networks (RNNs) and have proposed various fast RNN implementations~\cite{diamos2016persistent,skiprnn,zhang-sennrich-2019-lightweight}.
Notably, the Quasi-RNN~\cite{bradbury2016quasi} and SRU~\cite{lei2018sru} have invented highly-parallelizable recurrence and combined them with convolutions or highway networks respectively.
The resulting architectures achieve equivalent parallelism as convolutional and attention models.
This advancement eliminates the need of avoiding recurrence computation to trade model training efficiency, a design choice made by the Transformer architecture.
Our model builds on top of SRU.

Second, several recent works have argued that using attention alone is not the best architecture in terms of model expressiveness.
For example, \citet{dong2021} demonstrate theoretically and empirically that using pure attention results in performance degeneration.
\citet{gulati2020} have combined convolution and attention and obtained new state-of-the-art results for speech recognition.
Moreover, RNNs have been incorporated into Transformer architectures, resulting in improved results in machine translation and language understanding tasks~\cite{lei2018sru,huang2020transblstm}.
Our work is built upon a similar hypothesis that recurrence and attention are complementary at sequence modeling.
We demonstrate that jointly leveraging fast recurrence and attention not only achieves state-of-the-art modeling results but also obtain significant computation reduction.

Being orthogonal to our work, 
many recent works improve the efficiency of Transformer models by accelerating attention computation~\cite{zaheer2020bigbird,pmlr-v119-katharopoulos20a,vyas2020fast,peng2021rfa}.
Examples include Longformer~\cite{Beltagy2020Longformer}, Reformer~\cite{Kitaev2020Reformer}, Linformer~\cite{wang2020linformer} and Routing Transformer~\cite{roy2020efficient}.
In contrast, our work optimizes computational efficiency using recurrence combined with minimal attention and our model can incorporate these attention variants for additional speed improvement.

\section{Conclusion}
We present a highly-efficient architecture combining fast recurrence and attention,
and evaluate its effectiveness on various language modeling datasets.
We demonstrate fast RNNs with little attention not only achieve top results but also reduce training cost significantly.
Our work shares a different idea to accelerating attention, therefore providing an orthogonal direction to advancing state-of-the-art model architecture.
As future work, we believe the model can be improved using stronger attention or recurrent implementations, better normalization or optimization techniques.
\section*{Acknowledgement}
We would like to thank ASAPP Inc. for making this work possible.
We thank Hugh Perkins, Joshua Shapiro, Sam Bowman, Danqi Chen and Yu Zhang for providing invaluable feedback for this work.
Finally, we thank Jeremy Wohlwend, Jing Pan, Prashant Sridhar and Kyu Han for helpful discussions, and ASAPP Language Technology and Infra teams for the compute cluster setup for our research experiments.

\bibliography{emnlp2018}
\bibliographystyle{acl_natbib}

\clearpage
\newpage
\appendix


\section{Additional results}
\label{sec:appendix:more_analyses}

\subsection{Detailed analysis of attention}
Table~\ref{table:attn} presents a more comprehensive analysis of attention in SRU++ models.
First, we change the number of attention layers and their locations in the model.
As shown in the top block of Table~\ref{table:attn}, using attention in 50\% of the layers leads to no (or negligible) loss in model performance.
This is consistent with the results in Table~\ref{table:analyze_attention} using a smaller model.
Enabling attention in higher layers performs slightly better than evenly distributing attention from the bottom to top layers.

We also experiment with using more than one attention head in each of the attention layer, as shown in the middle block of the table.
Unlike Transformer models however, we do not observe a significant improvement using multiple heads.
We hypothesize that the recurrence states can already carry different features or information that are present in different input positions, making redundant heads unnecessary.

Finally, changing the ratio $d:d'$ from 4 to 8 gives similar improvements regardless of using 2 attention layers or 10 attention layers. 
This suggests that the amount of attention and the hidden size ratio can be tuned independently for best model performance.

\subsection{The effectiveness of layer normalization}
\label{sec:appendix:layer_norm}
\begin{figure*}[!t]
    \centering
    \includegraphics[width=6.2in]{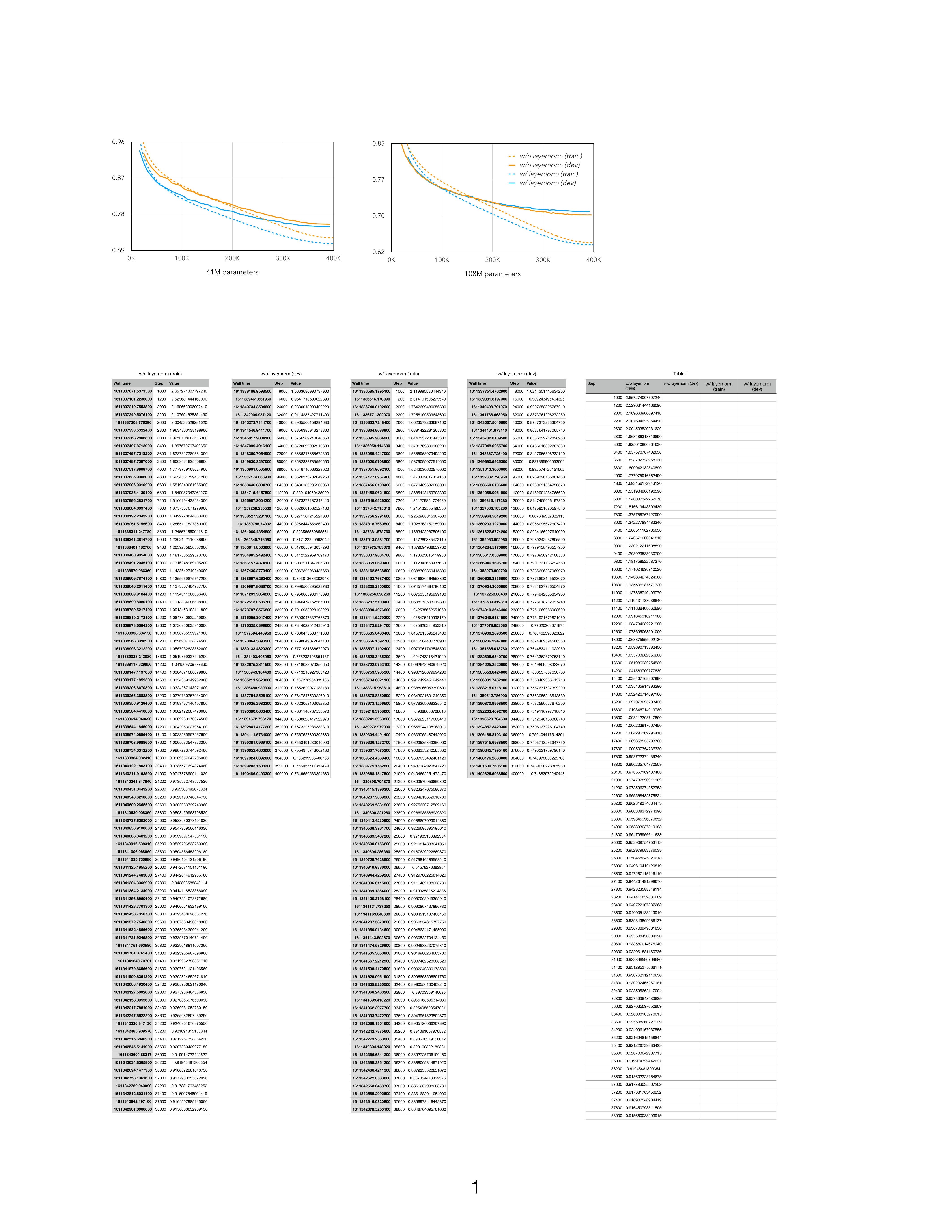}
    \vspace{0.15in}
    \caption{Understanding the empirical effect of layer normalization. We show the training and dev loss of SRU++ models using 41M parameters and 108M parameters on \textsc{enwik8} dataset. The model with layer normalization fits the training data better, but achieves worse generalization.}
    \label{fig:analyze_layernorm}
\end{figure*}

In our experiments, we have always used layer normalization to stabilize training.
However, we also found layer normalization to achieve worse generalization for larger models that are more prone to over-fitting.
Figure~\ref{fig:analyze_layernorm} showcases our empirical observation on the \textsc{enwik8} dataset. 
Using layer normalization achieves more rapid training progress and lower training loss, but results in higher dev loss in the case of training a 108M model.
This generalization gap remains even if we tune the dropout rate carefully.
In addition, although using layer normalization in the smaller model with 41M parameters gives slightly better dev results, we still observe a larger generalization gap (indicated by the difference between training loss and dev loss) compared to the run without layer normalization.
Similar over-fitting patterns are observed on \text{Wiki-103} dataset, and also in previous work~\cite{xu2019understanding}.

On the other hand, turning off layer normalization can achieve better generalization but makes training sensitive to learning rate and parameter initialization.
For example, we have to use a smaller learning rate of 0.00025 or lower to avoid sudden gradient explosion during training.
These results suggest possible future work by improving the normalization method~\cite{shen2020powernorm,brock2021characterizing}.

\subsection{Tuning weight decay and learning rate}
\label{sec:appendix:tuning_lr_wd}
We find that tuning the weight decay and learning rate critical to the success of training SRU++ and achieving best results.
Table~\ref{tab:lrwd} provides a sensitivity analysis by testing different learning rates and weight decay values. 
Increasing the weight decay consistently gives better results for all learning rates tested.
Tuning the learning rate is also needed to reach the best result.
The non-trivial effect of weight decay seems to be unique for SRU++.

On the other hand, the performance of SRU++ remains robust once the appropriate weight decay and learning rate are set. 
As shown in previous results and analyses, SRU++ achieves strong and relatively stable results to various hidden sizes, number of attention layers and datasets.
In particular, using the same weight decay value generalize well for all datasets (including language modeling and translation tasks) and model configurations tested.

\begin{table}[!h]
\centering
\begin{tabular}{r|ccc}
\hline
& {~~~}$0.10${~~~} & {~~~}$0.01${~~~} & {~~~}$0.00${~~~} \\
\hline
$3\times 10^{-4}\ $ & $\mathbf{1.014}$ & - & - \\
$2\times 10^{-4}\ $ & $1.022$ & $1.035$ & $1.047$ \\
$1.5\times 10^{-4}\ $ & $1.030$ & $1.038$ & $1.040$ \\
 \bottomrule
\end{tabular}
\caption{Dev BPC of SRU++ given a learning rate $\in \{1.5, 2, 3\} \times 10^{-4}$ and a weight decay $\in \{0.1, 0.01, 0\}$. `-` means the training run diverged or got gradient explosion.}
\label{tab:lrwd}
\end{table}

\section{Training details}
\label{sec:appendix:train_details}

\paragraph{Language modeling}
We use the RAdam optimizer\footnote{\url{https://github.com/LiyuanLucasLiu/RAdam}} with the default hyperparameters $\beta_1=0.9$ and $\beta_2=0.999$ for all our experiments.
We use a cosine learning rate schedule with only 1 cycle for simplicity.
For faster training, we also leverage the native automatic mixed precision (AMP) training and distributed data parallel (DDP) of Pytorch in all experiments, except those in Table~\ref{tab:enwik8_base} and Figure~\ref{fig:intro} for a fair comparison with the Transformer-XL implementation.

Table~\ref{tab:config_enwik8} shows the detailed training configuration of SRU++ models on \textsc{enwik8} dataset.
Most training options are kept the same for all models. 
We tune the dropout probability more carefully as we found training is more prone to over-fitting and under-fitting for this dataset.
The large model is trained with 2x batch size.
As a result, we increase the learning rate proportionally by a factor of $\sqrt{2}$ \cite{hoffer2017}, which results in a rounded learning rate of 0.0004.


Table~\ref{tab:config_wiki103} presents the detailed training configuration on \textsc{Wiki-103} dataset.
Similarly we use $d=3072$ and $d=4096$ for the base and large model respectively for a hidden size ratio $d:d' = 4:1$.
Following~\cite{baevski2018adaptive}, we use an adaptive word embedding layer and an adaptive softmax layer for our models, and we tie the weight matrices of the two layers.
We keep the total number of parameters comparable when we use a different hidden size ratio $d:d' = 8:1$.

\paragraph{Machine translation}
We use the open-sourced code from \citet{lin-etal-2020-autoregressive} for the \textsc{IWSLT}'14 De$\rightarrow$En translation task.
The Transformer model tuned by the original work uses 8 layers for both the encoder and decoder and a total of 20M parameters.
Most of the training configuration remains the same as the original work\footnote{\url{https://github.com/asappresearch/imitkd/blob/master/configs/iwslt/teacher.yaml}}, except for a couple of changes.
First, we use RAdam optimizer and the same $\beta$ values for consistency with the language model task.
We use the same weight decay value of 0.1 for SRU++.
The Transformer model uses a weight decay of 0 that is tuned based on dev set performance.
Second, we increase the number of training epochs to 50 (or equivalently 64K training steps) since all models achieve better BLEU scores by training longer.
This ensures we compare models when they reach the maximum performance.

Our SRU++ model uses a hidden size $d=1024$, an attention size $d'=256$ and 6 layers for the encoder and decoder, resulting in a similar number of parameters as the Transformer model in comparison.
Let $\mathbf{X}_{src}$ be the output representation of the SRU++ encoder.
Each SRU++ decoder layer make uses of $\mathbf{X}_{src}$ by simplying treating it as extra attention context. 
That is, the query, key and value representations are computed by concatenating the input of the current layer $\mathbf{X}_{tgt}$ with $\mathbf{X}_{src}$,
\begin{align*}
    \mathbf{Q} &= [\mathbf{Q}_{src}, \mathbf{Q}_{tgt}]\\
               &= \mathbf{W}^q\ [\mathbf{X}_{src}, \mathbf{X}_{tgt}]^\top \\
    \mathbf{K} &= \mathbf{W}^k\ \mathbf{Q} \\
    \mathbf{V} &= \mathbf{W}^v\ \mathbf{Q}
\end{align*}
The resulting representations $\mathbf{Q}_{tgt}$, $\mathbf{K}$ and $\mathbf{V}$ are used for the rest of the attention computation.
The attention mask is set such that each target token can only attend to all source tokens and preceding target tokens.

\begin{table*}[!t]
\centering
\begin{tabular}{lccc@{~~~~~~~~~~}cc}
\toprule
\bf Layers that has attention & \bf Num of heads & $d$ & $d'$ & \bf Model size & \bf Dev BPC \\
\midrule
All layers & 1 & 3072 & 768 & 108M & 0.997\\
6,7,8,9,10 & & & & 102M & 0.997\\
2,4,6,8,10 &  &  &  &  102M & 0.999\\
8,9,10 & & 3136 & 784 & 103M & 1.000\\
3,6,9 & & & & & 1.001\\
\midrule
5,10 & 1 & 3072 & 768 & 98M & 1.002\\
& 2 & & & & 1.002 \\
10 & 1 & &  & 97M & 1.007\\
& 2 & & & & 1.006\\
\midrule
All layers & 1 & 3072 & 768 & 108M & 0.997\\
5,10 & & & & 98M & 1.002\\
All layers & & 4480 & 560 & 109M & 0.991\\
5,10 & & & & 104M & 0.997\\
\bottomrule
\end{tabular}
\caption{Results of 10-layer SRU++ models by varying the attention setting.
We report the dev BPC on the \textsc{enwik8} dataset. The first column indicates layers where the attention are located. Smaller index numbers represent layers that are closer to the input of the model.}
\label{table:attn}
\end{table*}

\begin{table*}[!t]
\centering
\begin{tabular}{lcccc}
\hline
 & Base model & Base model & Large model & Large model \\
 & ($k=5$) & & & \\
\hline
Attention / unroll size - train & 1024 & 1024 & 1024 & 1024 \\
Attention / unroll size - test & 3072 & 3072 & 3072 & 3072 \\
Batch size $\times$ Num of GPUs & 4$\times$8 & 4$\times$8 & 8$\times$8 & 8$\times$8 \\
Dropout & 0.22 & 0.22 & 0.32 & 0.35 \\
Gradient clipping & 1.0 & 1.0 & 1.0 & 1.0 \\
Hidden size ratio $d:d'$ & 4 & 4 & 4 & 8 \\
Hidden size $d$ & 3072 & 3072 & 4096 & 6016\\
Hidden size $d'$ & 768 & 768 & 1024 & 752 \\
Learning rate & 0.0003 & 0.0003 & 0.0004 & 0.0004 \\
LR warmup steps & 16K & 16K & 16K & 16K \\
Training steps & 400K & 400K & 400K & 400K \\
Weight decay & 0.1 & 0.1 & 0.1 & 0.1 \\
\hline
Model size & 98M & 108M & 191M & 195M\\
Dev BPC & 1.002 & 0.997 & 0.985 & 0.974\\
Test BPC & 0.980 & 0.974 & 0.963 & 0.953\\
\bottomrule
\end{tabular}
\caption{Training details of SRU++ models on \textsc{enwik8} dataset.}
\label{tab:config_enwik8}
\vspace{0.8in}
\centering
\begin{tabular}{lcccc}
\hline
 & Base model & Large model & Large model & Large model \\
 & & & ($k=5$) & \\
\hline
Attention / unroll size - train & 768 & 1024 & 1024 & 1024\\
Attention / unroll size - test & 2560 & 2560 & 2560 & 2560\\
Batch size $\times$ Num of GPUs & 8$\times$8 & 8$\times$8 & 8$\times$8 & 8$\times$8 \\
Dropout & 0.15 & 0.2 & 0.2 & 0.2 \\
Gradient clipping & 1.0 & 1.0 & 1.0 & 1.0\\
Hidden size ratio $d:d'$ & 4 & 4 & 8 & 8 \\
Hidden size $d$ & 3072 & 4096 & 5952 & 5952\\
Hidden size $d'$ & 768 & 1024 & 744 & 744\\
Learning rate & 0.0003 & 0.0003 & 0.0003 & 0.0003\\
LR warmup steps & 16K & 16K & 16K & 16K\\
Training steps & 400K & 400K & 400K & 400K\\
Weight decay & 0.1 & 0.1 & 0.1 & 0.1\\
\hline
Model size & 148M & 232M & 225M & 234M\\
Dev PPL & 17.5 & 16.7 & 16.6 & 16.4\\
Test PPL & 18.3 & 17.4 & 17.3 & 17.1\\
\bottomrule
\end{tabular}
\caption{Training details of SRU++ models on \textsc{Wiki-103} dataset.}
\label{tab:config_wiki103}
\end{table*}

\end{document}